\documentclass{article}

\usepackage{lscape}
\usepackage{kao}
\usepackage[utf8]{inputenc}
\usepackage[french]{babel}

\title{Une architecture cognitive et affective \\ orient{\'e}e interaction}

\author{Damien Pellier, Carole Adam, Wafa Johal, \\ Humbert Fiorino and Sylvie Pesty\\[2mm]
UGA - LIG CNRS UMR 5217\\
Grenoble, France}

\date{13-14 juin 2016}

\begin{document}

\maketitle

\begin{abstract}
Les robots trouvent de nouvelles applications dans notre vie de tous les jours et interagissent de plus en plus étroitement avec leurs utilisateurs humains. Cependant, malgré une longue tradition de recherche, les architectures cognitives existantes restent souvent trop génériques et pas assez adaptées aux besoins spécifiques de l'Interaction sociale Humain-Robot, comme la gestion des émotions, du langage, des normes sociales, etc. Dans cet article, nous présentons CAIO, une architecture Cognitive et Affective Orientée Interaction. Elle permet aux robots de raisonner sur les états mentaux (y compris les émotions) et d'agir physiquement, émotionnellement et verbalement.
\end{abstract}

\section{Introduction}

Les robots sont de plus en plus présents dans notre vie quotidienne, dans des rôles d'assistants, de robots pédagogiques, de compagnons pour les enfants ou les personnes âgées, etc. rôles qui impliquent de nombreuses interactions avec l'utilisateur. Cette proximité signifie que les robots doivent partager non seulement le même espace physique mais aussi des buts et des croyances pour accomplir une tâche commune. Dans l'idéal, ils devraient pouvoir interagir en langage naturel mais aussi non-verbalement par gestes et expressions faciales.
De nombreuses applications qui relèvent du domaine de l'Interaction Humain-Robot (HRI - Human-Robot Interaction) sont en cours de développement, mais l'architecture cognitive sous-jacente du robot n'est pas toujours explicite. En effet, le développement d'une architecture cognitive permettant à un robot de gérer la complexité des interactions avec des humains est un  défi en soi. De nombreuses fonctionnalités sont nécessaires: gestion de la mémoire, raisonnement sur les états mentaux, gestion du dialogue, gestion des émotions et des aspects non-verbaux de l'interaction etc.


Malgré les différentes contributions dans le domaine des architectures cognitives et celles plus spécifiques pour la robotique \cite{pattachi:12,lemaignan11}, la plupart des architectures sont génériques et peu d'entre elles peuvent vraiment gérer la complexité des interactions Humain-Robot.
Dans cet article nous présentons notre Architecture Cognitive et Affective Orientée Interaction (CAIO) dont les contributions se situent sur les aspects suivants~: autoriser le robot à raisonner sur les états mentaux (incluant les émotions) et à expliquer ses actions en termes d'états mentaux~; gérer à la fois les actions physiques et verbales~; et fournir deux boucles de réponse distinctes (réactive et délibérative).

Cet article est structuré comme suit. La section~\ref{S:RelatedWorks} présente quelques travaux sur les architectures cognitives et robotiques. La section~\ref{previous} revient brièvement sur nos précédents travaux à la base de  l'architecture CAIO que nous détaillons dans la suivante (section~\ref{caio}). La section~\ref{eval} donne ensuite un court exemple sur un robot Nao, et nous positionnerons l'architecture CAIO en nous basant sur des critères définies par Langley et al. \cite{Langley2009a}. La Section~\ref{conclu} donne quelques perspectives.

\section{Etat de l'art}          
\label{S:RelatedWorks}           

Les travaux dans le domaine des architectures cognitives sont une recherche de longue date et des états de l'art peuvent être trouvés dans différents articles, par exemple \cite{Thorisson2012,Chong2009}. Trois grandes familles d'architectures cognitives sont généralement distinguées: (1) les architectures dites bio-inspirées dont l'ambition est principalement la compréhension des processus de la cognition, (2) les architectures qualifiées de "purement" Intelligence Artificielle et (3) les architectures qui trouvent leur fondement dans des travaux en philosophie (la philosophie de l'esprit essentiellement). Nous donnons ici quelques éléments sur les principales architectures.

\subsection{Les Architectures bio-inspirées}

ACT-R (Adaptive Control of Thought-Rational) est une des architectures les plus connues pour modéliser les mécanismes de la cognition humaine \cite{Anderson2005} \cite{AnBo1973}. Elle repose principalement sur la séparation entre deux types de connaissance: la connaissance déclarative (chunks) et la connaissance procédurale (rules). L'architecture ACT-R est composée de 4 principaux modules, chacun ayant un équivalent chez l'humain (visual module - perception, goal module, declarative memory, and manual module - actuators), dont la coordination est assurée par un système central.

L'architecture CLARION (Connectionist Learning with Adaptive Rule Induction ON-line) \cite{Sun2002} est également très connue; c'est une architecture hybride qui combine des représentations connexionnistes et des représentations symboliques, et combine également des processus psychologiques implicites et explicites. L'ambition des recherches autour de CLARION est principalement de comprendre les mécanismes de la cognition humaine (prise de décision, apprentissage raisonnement, etc.) tout en développant également des agents artificiels.

ASMO (Attentive Self-MOdifying) \cite{NoJoWi2010} a été développée plus récemment à partir d'une théorie biologique de l'attention, pour résoudre les compétitions entre buts éventuellement incompatibles du robot. Concrètement, le niveau d'attention détermine les priorités relatives des buts, les plus critiques étant traités comme des réflexes. ASMO a été implémentée dans un robot social en interaction avec des humains, et dans un robot Nao pour des compétitions de football.

\subsection{Architectures pour la résolution de problèmes d'Intelligence Artificielle}

SOAR (State, Operator And Result) \cite{Laird2012} est une architecture purement "IA symbolique", qui met l'accent sur l'apprentissage et la résolution de problèmes. Elle dispose  d'une mémoire à court-terme (ou mémoire de travail), et une mémoire à long-terme qui se décompose en mémoire procédurale, mémoire sémantique et mémoire épisodique. L'apprentissage par renforcement est déclenché quand les connaissances ne permettent pas de prendre une décision. SOAR a été étendue avec des émotions qui interviennent au niveau de l'apprentissage \cite{soar-emo}.

L'approche choisie pour l'architecture ICARUS \cite{LaChRo2005} consiste à unifier les techniques réactives et délibératives de résolution de problèmes, ainsi que le raisonnement numérique et symbolique. Un but non satisfait de plus haute priorité reçoit l'attention: les capacités connues permettant de le réaliser sont amenées de la mémoire à long-terme vers la mémoire à court-terme, ou bien ce but est décomposé en sous-buts pour apprendre de nouvelles capacités.

\subsection{Architectures basées sur des théories philosophiques}

La philosophie de l'esprit s'intéresse à l'ensemble des problèmes qui se posent dans la vie mentale et en particulier pour l'action, la perception, le raisonnement, l'intentionnalité. En philosophie de l'esprit, il y a 3 états mentaux de base : croire et désirer et l'intention d'agir.
La philosophie de l'action est une branche de la philosophie de l'esprit et s'intéresse plus particulièrement à l'{\em action}.  La théorie de Bratman de l'action planifiée (Human Practical reasoning) \cite{Bratman87} a ainsi servi aux travaux sur les logiques et les agents BDI ({\em Belief, Desire, Intention}) \cite{RaoGeorgeff91,CohenLevesque}). Dans cette théorie, les croyances et les désirs sont la cause de l'intention d'action;
L'architecture PRS pour les agents rationnels (\eg PRS, {\em Procedural Reasoning System} \cite{wooldridge}) est une des architectures très connues. Comme nous l'expliquerons plus loin, l'architecture CAIO s'inscrit dans cette lignée, mais avec l'addition de nouveaux états mentaux (ou attitudes propositionnelles), en particulier les {\em émotions} qui sont essentielles pour un robot expressif social,
à cause du rôle majeur qu'elles jouent dans l'interaction et le raisonnement.

Actuellement, plusieurs recherches se tournent vers la Théorie de l'esprit (Theory of Mind - ToM) \cite{sabouret-tom1, sabouret-tom3, Warnier2012} qui est également une branche de la philosophie de l'esprit. Cette théorie s'occupe de la compréhension d'autrui et de soi. IL s'agit de comprendre ce que les autres vont faire, d'être capable de le prédire afin d'anticiper les conséquences de ce comportement.

\section{Précédents travaux} \label{previous}

\subsection{La logique BIGRE pour représenter les états mentaux et les émotions}

Guiraud et al. \cite{guiraud11} ont proposé une logique modale dans la lignée des logiques BDI (Belief, Desire, Intention), pour représenter formellement les états mentaux qu'un agent artificiel (un robot par exemple) va pouvoir exprimer lors de son interaction avec un humain.

Cette logique, dite logique BIGRE (Belief, Ideal, Goal, Responsibility, Emotion), définit 5 états mentaux \footnote{Le lecteur est invité à se référer à \cite{guiraud11} pour plus de détails sur la sémantique et l'axiomatique de cette logique.}:
\begin{itemize}
\item \emph{Belief} (B) $Bel_{i}\varphi$: le robot $i$ croit $\varphi$ vrai;
\item \emph{Ideal} (I) $Ideal_{i}\varphi$: idéalement pour le robot $i$, $\varphi$ devrait être vrai (normes morales et sociales du robot \footnote{Par exemple, une obligation morale à aider une personne en danger ou une règle sociale à payer les taxes, etc.});
\item \emph{Goal} (G) $Goal_{i}\varphi$: le robot $i$ a pour but que $\varphi$ soit vrai;
\item \emph{Responsibility} (R) $Resp_{i}\varphi$: le robot $i$ est responsable de $\varphi$ ;
\item \emph{Complex emotion} (E) (\eg gratitude, admiration, reproche, etc.) qui résulte d'un raisonnement à propos de \emph{Responsibility}. \end{itemize}

Les émotions complexes dont il est question ici jouent un rôle central dans l’interaction dialogique et sont exprimées principalement par le langage.
Ce sont des émotions qu'il faut distinguer des émotions de base (voir ci-dessous) que l'on peut représenter à partir des \emph{Belief}, des \emph{Goal} et des \emph{Ideal} qui sont quant à elles associées généralement à des expressions faciales prototypiques.

Les opérateurs B, I, G et R  (BIGR $\rightarrow$ E) ont permis de représenter 12 émotions: 4 émotions de base ({\em joie}, {\em tristesse}, {\em approbation}, {\em désapprobation}) et 8 émotions complexes ({\em regret}, {\em déception}, {\em honte}, {\em reproche}, {\em satisfaction morale}, {\em admiration}, {\em réjouissance} et {\em gratitude}, voir tableau \ref{Tab:ComplexEmotions}).

La logique BIGRE a de plus permis de définir formellement des actes de conversation expressifs (regretter, reprocher, complimenter, etc.), actes qui expriment des émotions complexes (le regret, le reproche, l'approbation morale, etc.). (voir section suivante).

\begin{table*}
\begin{center}
\begin{tabular}{|c||ccc|}
\hline
$\wedge$ & $Bel_{i\varphi}$ & $Bel_{i}Resp_{i\varphi}$ & $Bel_{i}Resp_{j\varphi}$ \\
\hline \hline
$Goal_{i\varphi}$ & $Joy_{i\varphi}$ & $Rejoicing_{i\varphi}$ & $Gratitude_{i,j\varphi}$ \\
$Goal_{i\neg\varphi}$ & $Sadness_{i\varphi}$ & $Regret_{i\varphi}$ & $Desappointment_{i,j\varphi}$ \\
$Ideal_{i\varphi}$ & $Approval_{i\varphi}$ & $MoralSatisfaction_{i\varphi}$ & $Admiration_{i,j\varphi}$ \\
$Ideal_{i\neg\varphi}$ & $Disapproval_{i\varphi}$ & $Guilt_{i\varphi}$ & $Reproach_{i,j\varphi}$ \\
\hline
\end{tabular}
\end{center}
\caption{Formalisation des 4 émotions de base et des 8 émotions complexes.}  \label{Tab:ComplexEmotions}
\end{table*}

\begin{table*}[!]
\centering
\begin{tabular}{|l|p{8cm}|}
\hline
\textit{Rejoice} & $Exp_{a,h,H}(Rejoicing_{a}\varphi)$\\
					& $\equiv  Exp_{a,h,H}(Goal_{a}\varphi \wedge Bel_{a}Resp_{a} \varphi)$\\
\hline
\hline
Preconditions &   $Goal_{a}\varphi \wedge  Bel_{a}Resp_{a} \varphi \eqdef Rejoicing_{a}\varphi$  \\
 &	Agent \textit{a} \guill{feels} rejoicing; it believes it is responsible for having achieved its goal $\varphi$.\\
\hline
Sending effects &  $Bel_{a}Bel_{h}Rejoicing_{a}\varphi$  \\
 &	Agent \textit{a} believes that human \textit{h} believes that \textit{a} is rejoicing about $\varphi$.\\
\hline
Reception effects  & $Bel_{a}Goal_{h}\varphi \wedge Bel_{a}Bel_{h}Resp_{h}\varphi $ \\
 &	Agent \textit{a} believes that human \textit{h} expressed his rejoicing about $\varphi$. Therefore, \textit{a} believes that \textit{h} has achieved its goal $\varphi$ and believes himself to be responsible for this.\\
\hline
\end{tabular}
\caption{Exemple: l'acte ACM \textit{Se réjouir - Rejoice} du point de vue de l'agent \textit{a} en conversation avec un humain \textit{h}.}
\label{tab:rejouir}
\end{table*}

\subsection{Le Langage de Conversation Multimodal}

Pour qu'un agent artificiel, type personnage virtuel, puisse s'exprimer de manière crédible, Riviere \etal \cite{riviere2011} ont défini un langage de conversation pour agent basé sur la Théorie des Actes de Discours de Searle \cite{searle85}.
Ce langage est appelé {\em Langage de Conversation Multimodal (LCM)} car sa particularité est de relier étroitement les aspects verbaux (les énoncés) aux aspects non-verbaux (les expressions faciales/gestes) afin d'augmenter l'expressivité de l'agent. Ce langage est dans la lignée des langages de communication pour agent tels que FIPA \cite{fipa}.
\\

Ce langage LCM regroupe 38 \emph{Actes de Conversation Multimodaux} (ACM) répartis en 4 classes:

\begin{itemize}
\item actes assertifs: \emph{dire}, \emph{affirmer}, \emph{nier}, etc.
\item actes directifs: \emph{demander}, \emph{suggérer}, \emph{réclamer}, etc.
\item actes engageants: \emph{promettre}, \emph{accepter}, \emph{refuser}, etc.
\item actes expressifs: \emph{remercier}, \emph{reprocher}, \emph{féliciter}, etc.
\end{itemize}

\noindent Chaque acte ACM a été formalisé en logique BIGRE :
\begin{itemize}
\item préconditions de l'acte, que l'agent doit satisfaire avant de pouvoir émettre cet acte, ceci pour garantir la {\em sincérité} de l'agent au sens de la théorie de Searle (sincerity conditions); 
\item effets en émission de l'acte, pour l'agent artificiel qui s'exprime;
\item effets en réception de l'acte, pour l'agent artificiel qui reçoit l'acte d'un interlocuteur.
\end{itemize}

\noindent Par exemple, le tableau~\ref{tab:rejouir} donne la formalisation de l'acte ACM \textit{Se réjouir - Rejoice}.
%

Cette représentation formelle permet à l'agent de mettre à jour ses états mentaux, en particulier ses croyances mais aussi ses émotions, les états mentaux de son interlocuteur, et de savoir s'il lui est possible de s'exprimer.

\subsection{Le moteur de raisonnement PLEIAD}

PLEIAD (ProLog Emotional Intelligent Agent Designer) \cite{Adam2007} est à l'origine un moteur de raisonnement pour des agents de type BDI, développé en SWI-Prolog. Il confère aux agents des capacités génériques de raisonnement. Il permet également l'implémentation de différents modèles d'émotions, comme le modèle OCC \cite{AHL2009}, ou de théories sur des émotions particulières comme la honte \cite{AdLon2013}.
PLEIAD a été plus récemment complété avec des stratégies de \emph{coping} \cite{pleiad2014} et des traits de personnalité.

Le moteur de raisonnement PLEIAD a été utilisé pour implémenter la logique BIGRE et constitue le coeur de l'architecture CAIO, principalement pour le module \emph{Deliberation} et pour la partie cognitive du module \emph{Evaluation des émotions} (voir section~\ref{caio:modules}).

\section{L'architecture CAIO} \label{caio}
\label{Moteur}


\subsection{Vue globale de l'architecture}

\begin{figure}[!t]  
\centering
\includegraphics[width=0.9\columnwidth]{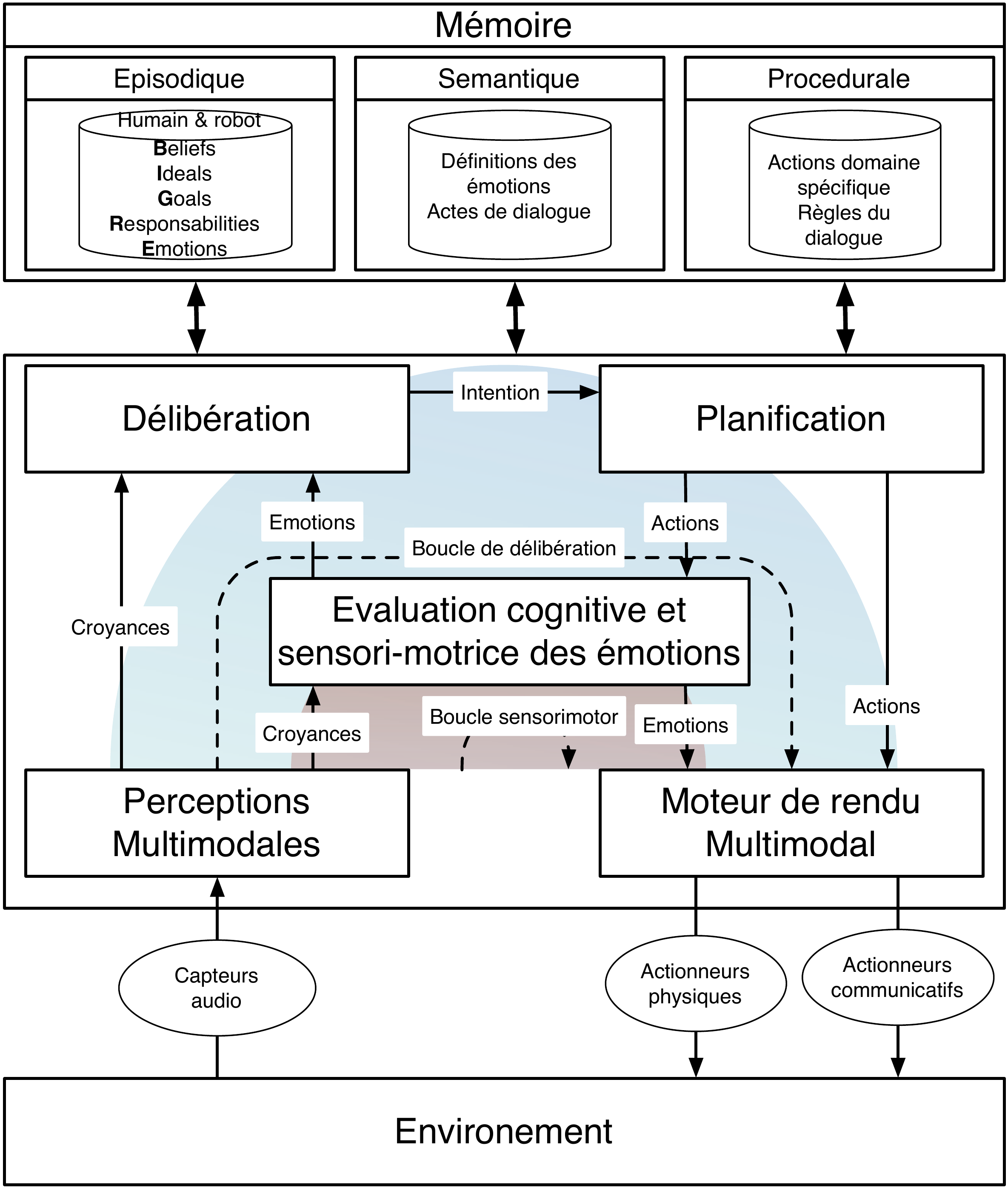}
\caption{Schéma conceptuel de l'architecture CAIO.}
\label{fig:archiCog}
\end{figure}

Une des spécificités de l'architecture CAIO (cf.~Figure~\ref{fig:archiCog}) est d'avoir 2 boucles de raisonnement: {\it une boucle délibérative} qui permet de raisonner sur les états mentaux BIGRE et de produire des plans d'actions au sens de la planification automatique et {\it une boucle sensori-motrice} plus rapide qui permet d'exprimer de manière réactive les émotions. Chaque boucle prend en entrée les perceptions multimodales en provenance de l'environnement.


Au cours de la boucle délibérative, la partie cognitive du module d'évaluation des émotions déduit les émotions complexes à partir des états mentaux. 
Le module de délibération déduit alors les {\it intentions} du robot à partir de ses états mentaux et choisit parmi celles-ci la plus appropriée. A partir de cette intention, le module de planification produit un plan, i.e., un ensemble ordonné d'actions communicatives ou physiques, pour atteindre l'intention retenue et choisir la prochaine action à exécuter. Finalement, le moteur de rendu multimodal exécute l'action précédemment choisie. Les modules de l'architecture fournissent un retour aux autres modules sur le résultat de leur traitement. Par exemple, le module de planification  informe le module de délibération lorsque l'intention choisie n'est pas atteignable ou encore le moteur de rendu multimodal informe le planificateur du succès ou de l'échec de l'exécution d'une action.


Simultanément, au cours de la boucle sensori-motrice, le module d'évaluation des émotions évalue les données provenant du module de perception multimodale selon les critères de Scherer (critères SEC - Stimulus Evaluation Checks).Ce processus d'évaluation détermine le déclenchement des émotions et envoie au module de rendu multimodal les actions faciales et gestuelles à exécuter pour exprimer l'émotion.


\subsection{Principaux modules de l'architecture}
\label{caio:modules}


\subsubsection{Le module de perception multimodale}

Le rôle de ce module est de fusionner les informations en provenance des capteurs du robot et de les exprimer sous la forme d'états mentaux du locuteur. Pour l'instant, nous nous sommes focalisés essentiellement sur les aspects associés au langage. Néanmoins, nous envisageons à court terme d'intégrer d'autres données en entrée telles que les expressions faciales ou encore les signaux para-linguistiques.


Concrètement, ce module effectue dans un premier temps un traitement de la parole en utilisant l'interface de Google Speech. A partir de l'énoncé textuel, le module, dans un second temps, extrait l'acte de conversation ACM. \footnote{le traitement de la langue est un champ de recherche complexe; nous n'abordons pas cet aspect dans ce travail, l'extraction des actes ACM à partir des énoncés est faite de façon ad-hoc.}.


\subsubsection{La mémoire}
\label{SS:BeliefRevisionModule}

La mémoire au sein de l'architecture CAIO est divisée classiquement en 3 parties. La mémoire \textit{épisodique} contient d'une part les états mentaux BIGRE propre au robot et d'autre part ceux de l'humain. La mémoire \textit{sémantique} contient la définition des émotions et des actes de conversation. La mémoire \textit{procédurale} gère la description des actions physiques que peut réaliser le robot et les règles du discours (cf. Tableau~\ref{Tab:Obligation-Rules}).


La mémoire est dynamiquement mise à jour en trois temps. Tout d'abord, les nouvelles croyances déduites à partir des perceptions du monde ou résultantes des interactions, i.e., les effets des actes ACM du locuteur et les effets produits par les propres actes ACM du robot, sont ajoutées à la mémoire. Puis les états mentaux sont mis à jour par inférence.




\subsubsection{Le module d'évaluation des émotions} \label{SS:Appraisal}

Le module d'évaluation est constitué de deux sous-modules: l'évaluation cognitive et l'évaluation sensori-motrice:
\begin{itemize}
\item l'évaluation cognitive prend en entrée les perceptions du robot ainsi que ses états mentaux et calcule l'émotion correspondante à partir de leur définition logique (cf.~tableau~\ref{Tab:ComplexEmotions}). Par exemple, une émotion de gratitude est calculée, lorsque le robot a le but $\varphi$ et croit que l'humain est responsable de ce but, i.e., lorsque le robot \textit{i} à l'état mental $\goal{i}\varphi \wedge \bel{i}\resp{j} \varphi$. Une intensité pour l'émotion est dérivée à partir des priorités du but ou de l'idéal précisé dans sa définition. Cette émotion est stockée dans la mémoire épisodique.

\item L'évaluation sensori-motrice évalue tous les actes ACM reçus du locuteur ou ceux émis par le robot lui-même selon des critères dévaluation définis par Scherer (\emph{SEC-Stimulus Evaluation Checks})  \cite{scherer2001} que nous avons adaptés dans le cadre de nos travaux:
\begin{itemize} 
  \item la \textit{nouveauté} de l'acte de conversation (est-ce que l'acte était prévu dans le schéma général du dialogue ?);
	\item le \textit{plaisir intrinsèque} (en fonction du type d'acte, e.g., \emph{Refuse} \textit{vs.} \emph{Accept}, et de son contenu propositionnel);
	\item la \textit{congruence} rapport aux buts
	\item le \textit{potentiel d'adaptation} (est-ce que le robot peut influencer les conséquences des actes ?);
	\item la \textit{compatibilité} avec les idéaux du robot et les normes sociales.
\end{itemize}


Le résultat de cette évaluation sensori-motrice  est propagé au module de rendu afin d'être traduit en expressions faciales et corporelles. (cf.~tableau~\ref{tab:secfear}).

\end{itemize}


\subsubsection{Le module de délibération}
\label{SS:Deliberation}

La délibération est le processus qui permet au robot de décider de sa prochaine {\it Intention}  (ou but). Nous définissons trois types d'{\it Intention}.


Les {\em Intentions émotionnelles} sont les intentions qui vont se traduire en général par un acte ACM expressif (par exemple regretter, remercier, reprocher, se réjouir,...). Afin que le robot soit sincère, affectif et expressif, nous avons décidé que toute émotion "ressentie" par le robot devait être exprimée, ce qui aide à une meilleure régulation locale du dialogue. \cite{Bates94}. Notons que les intentions émotionnelles ont la priorité la plus haute.


Les {\em Intentions fondées sur les obligations} contribuent aussi à la régulation du dialogue et conduisent à des interactions humain-robot plus naturelles \cite{Baker}. Les intentions de ce type sont adoptées à partir des {\em règles du discours} définies par Traum et Allen  \cite{TraumAllen} pour représenter les normes sociales et guider le comportement du robot. Ces intentions jouent un rôle important dans la réaction du robot au niveau du dialogue (cf.~tableau~\ref{Tab:Obligation-Rules}). Concrètement le robot adopte toujours l'intention d'atteindre ses obligations (déduites à partir des règles du discours) mais contrairement aux intentions émotionnelles, il leur affecte une priorité moyenne.


\begin{table}
\begin{center}
\begin{tabular}{lp{3.5cm}}
{\bf Origine de l'obligation } & {\bf Action à réaliser} \\
\hline \hline
$S_1$ Accept ou Promise $A$ & $S_1$ achieves $A$ \\
$S_1$ Request $A$ & $S_2$ address Request: accept $A$ {\bf ou} reject $A$ \\
$S_1$ Question fermée est-ce que $P$ & $S_2$ Answer-If $P$ \\
$S_1$ Question ouverte $P(x)$ & $S_2$ Inform-ref $x$ \\
\end{tabular}
\end{center}
\caption{Exemples de règles du discours définies par Traum et Allen entre deux locuteurs (S1) et (S2) portant sur une proposition (P).}
\label{Tab:Obligation-Rules}
\end{table}

Finalement, l'{\em Intention globale} exprime la direction générale du dialogue et définit son type \cite{walton95commitment}, e.g., délibération, persuasion, etc. Ce type d'intention est adoptée lorsque le robot s'engage à l'atteindre, soit explicitement, e.g., en l'explicitant par un acte ACM de type \emph{Promettre} ou \emph{Accepter}, ou en interne en raisonnant sur ses croyances. L'intention globale a toujours la priorité la plus faible.


\subsubsection{Le module de planification}

Le module de planification est en charge de trouver un moyen d'atteindre l'Intention décidée par le module de délibération. Ce module repose sur l'idée proposée à l'origine par \cite{PerraultAllen} que l'ordonnancement des actes de dialogue peut être réalisé par la planification. En termes d'implémentation, le module repose sur la librairie PDDL4J \cite{pellier:15} et les travaux de \cite{pellier:14}. Il est important de noter que les plans produits peuvent contenir à la fois des actes de conversation ACM et des actions physiques, qui peuvent tous les deux être exécutés par le robot. Les pré-conditions  et les effets des actions sont formalisés dans le langage PDDL ({\it Planning Domain Description Language}) utilisé dans la communauté planification rendant ainsi l'architecture CAIO compatible avec la plupart des planificateurs.


Dans le cas des Intentions émotionnelles et de celles fondées sur les obligations, le plan généré est souvent constitué d'un seul acte de conversation ACM. Par exemple, l'intention émotionnelle pour exprimer la gratitude peut être atteinte avec un acte de conversation ACM \textit{Remercier} ou \textit{Féliciter} en fonction de l'intensité de l'émotion.

Dans le cas des autres intentions, il est nécessaire de mettre en {\oe}uvre des actions spécifiques au domaine dont les préconditions et les effets sont décrits dans la mémoire procédurale du robot. Par exemple, pour réserver des billets de train, il est nécessaire de connaître l'heure et la date de départ ainsi que le lieu de départ et la destination. Le planificateur peut donc être amené dans ce cas à produire un plan contenant à la fois des actes de conversation ACM, e.g., demander les informations pertinentes à l'utilisateur, et des actions spécifiques au domaine, e.g., réserver des billets de train.

%

Si aucun plan ne peut être généré pour atteindre l'intention déterminée par le module de délibération, l'intention est abandonnée et le module de délibération est une nouvelle fois sollicité pour calculer une nouvelle intention.


\subsubsection{Le moteur de rendu multimodal} \label{renderer}

Le moteur de rendu multimodal prend en entrée la première action orchestrée par le module de planification et l'émotion produite par le module d'évaluation des émotions. Puis, il agit sur les actionneurs du robot pour offrir un rendu physique de l'action intégrant dynamiquement l'expression faciale et les postures correspondant à l'émotion. Comme mentionné précédemment, la génération des postures s'appuie sur les valeurs des critères d'évaluation SEC \cite{scherer2001,erden2013,coulson2008} calculées par le sous-module sensori-moteur du module d'évaluation des émotions, par exemple le tableau~\ref{tab:secfear} montre les postures produites pour exprimer le reproche.



\begin{landscape}
\begin{table}[htb]
  \centering
  \vspace{3cm}
		\begin{tabular}{p{2.5cm}p{0.2cm}p{2.5cm}p{0.2cm}p{2.5cm}p{0.2cm}p{2.5cm}p{0.2cm}p{2.5cm}}
    Nouveauté & $\Rightarrow$&
		Plaisir \hspace{1,5cm}intrinsèque &$\Rightarrow$&
		Effet sur \hspace{1cm} But/Besoin  & $\Rightarrow$&
		Potentiel d'adaptation &$\Rightarrow$&
		Norme \hspace{1cm}Compatibilité \\
			\includegraphics[height=3cm]{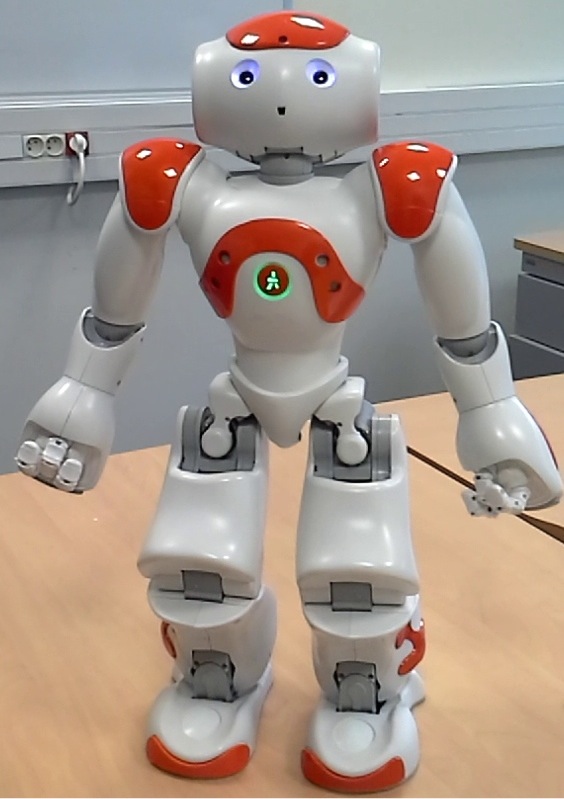} &$\Rightarrow$&
      \includegraphics[height=3cm]{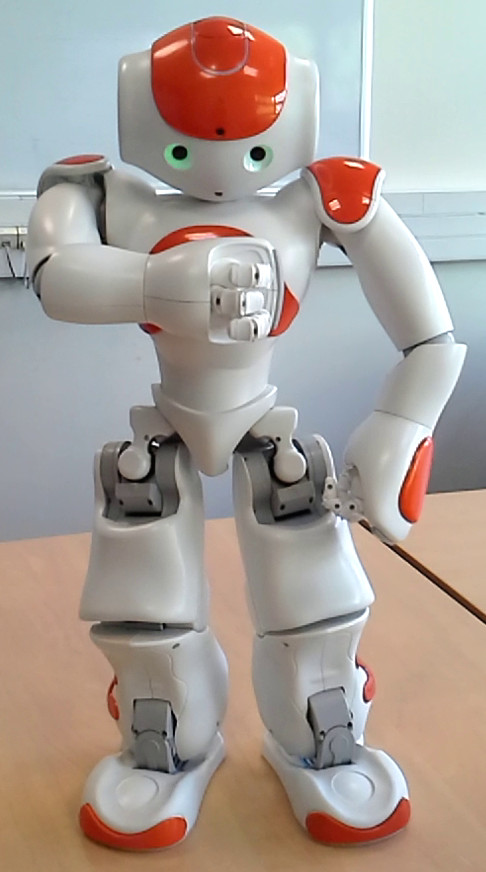}&$\Rightarrow$&
			\includegraphics[height=3cm]{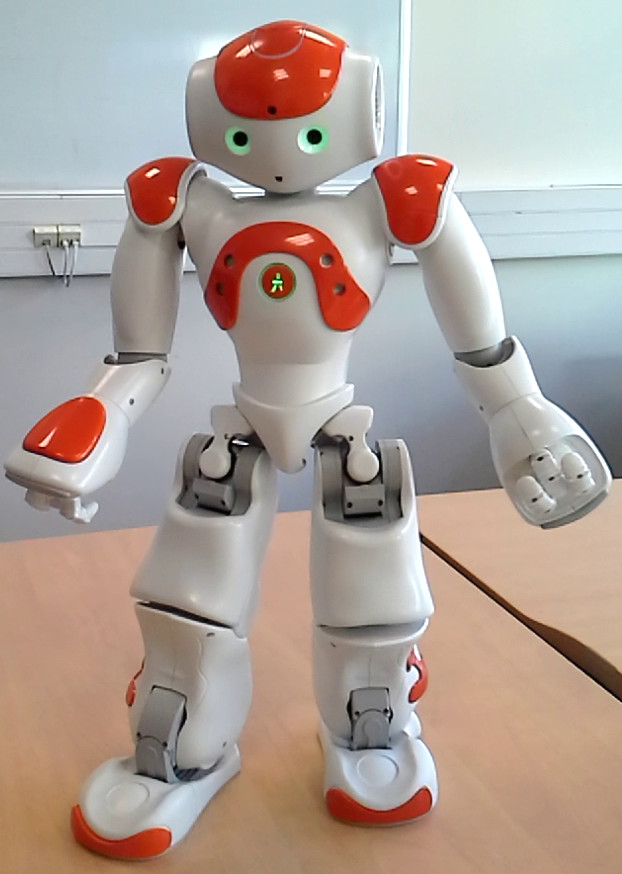}& $\Rightarrow$&
			\includegraphics[height=3cm]{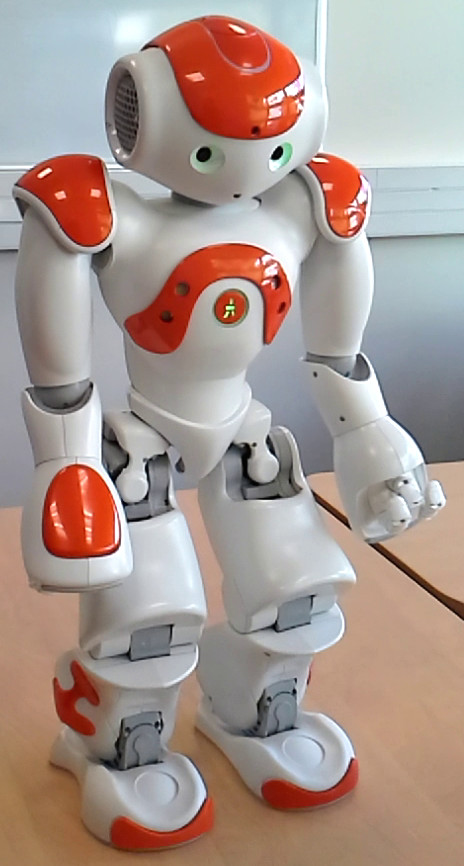}& $\Rightarrow$&
			\includegraphics[height=3cm]{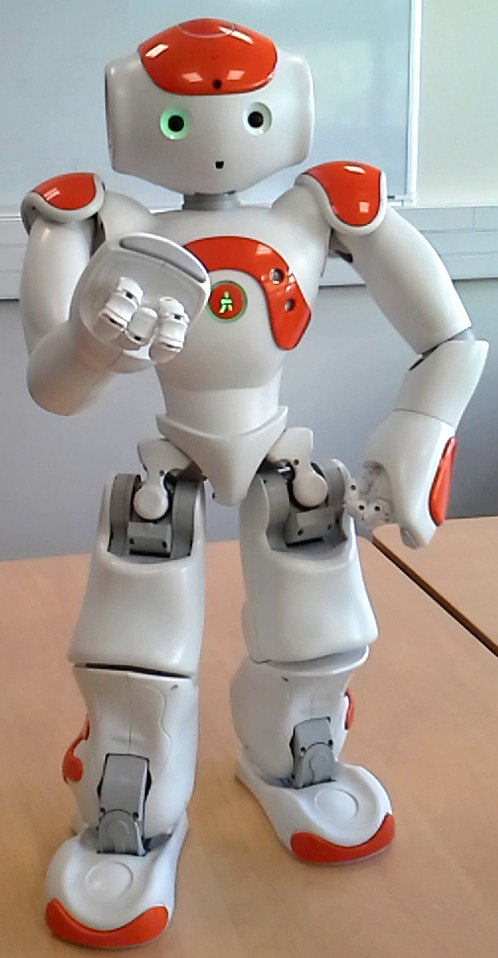} \\
			Nouveau & &
			Déplaisant &&
			Attentes insatisfaites  & &
			Peu de contrôle &&
			Norme violée \\
		\end{tabular}
	\caption{Séquence d'évaluation sensori-motrice qui exprime le reproche.}
	\label{tab:secfear}
\end{table}
\end{landscape}

\section{Exemple et évaluation}                            
\label{eval}                                    

\subsection{Exemple pour le robot Nao}

Une version préliminaire de l'architecture CAIO a été implémentée pour un personnage virtuel afin d'évaluer la crédibilité des comportements produits par l'architecture. Elle a notamment été testée dans le cadre d'un scénario de dispute entre un personnage virtuel compagnon et son utilisateur \cite{riviere2012}. Cette première version était conçue pour un personnage virtuel et n'était pas modulable. Une nouvelle version de l'architecture CAIO a été développée avec ROS (Robot Operating System). ROS est utilisé et largement répandu dans la communauté robotique. Les modules sont principalement implémentés en python et interfacés pour certains d'entre eux: avec SWI-Prolog pour le module de délibération et avec Java pour le module de planification PDDL4J.



\begin{figure}[t]
\centering
\includegraphics[width=\linewidth]{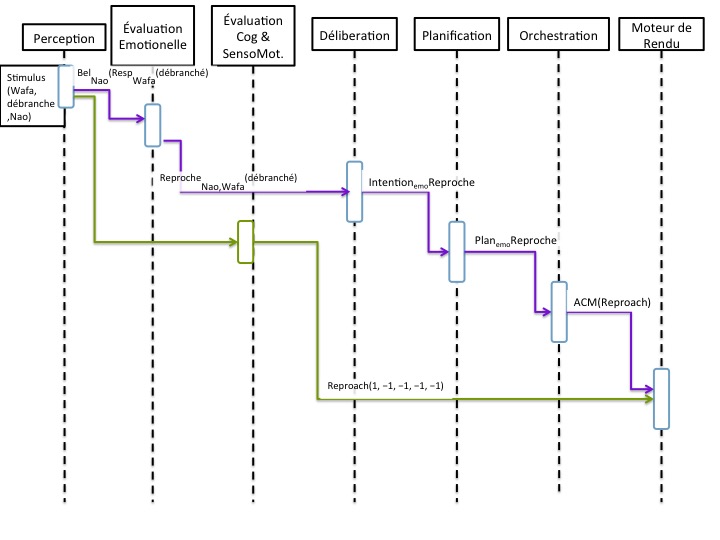}
\caption{Diagramme de séquence UML}
\label{fig:CAIOBasicSequenceDiagram}
\end{figure}

Nous présentons ci-dessous un court scénario illustrant l'encapsulation des différents modules de l'architecture CAIO dans des n{\oe}uds ROS (cf. le diagramme de séquence UML de la figure~\ref{fig:CAIOBasicSequenceDiagram})


Ce scénario implique deux acteurs: un utilisateur, \textit{Wafa}, et le \textit{robot Nao}. Nao a une batterie en fin de vie qui nécessite d'être branchée en permanence. Dans sa mémoire épisodique, Nao possède l'Idéal d'être branché exprimé de la manière suivante: $Ideal_{Nao}(\neg unplugged)$. De son côté Wafa doit sortir ce soir et doit se sécher les cheveux, mais Nao est branché à la seule prise disponible près du miroir. Par conséquent, Wafa annonce à Nao: ''~Nao, je vais te débrancher, j'ai besoin de me sécher les cheveux~''.


\begin{enumerate} 
\item Le n{\oe}ud \textit{perception} reçoit le premier acte ACM initialisant le dialogue suivant: \petit{$Stimulus(unplugged, wafa, nao)$} ;
\item Cette acte est traduit en termes d'états mentaux comme suit: \petit{$Bel_{Nao}(unplugged)$} et \petit{$Bel_{Nao}(Resp_{Wafa}(unplugged))$} ;
\item le module d'évaluation des émotions (sous-module cognitif) déduit l'émotion \petit{$Reproach_{Nao,Wafa}(unplugged)$} à partir des états mentaux;
\item Le module d'évaluation des émotions (sous-module sensori-moteur) évalue l'acte selon les 5 critères d'évaluation ; 
\item Le n{\oe}ud \textit{déliberation} reçoit l'émotion de \emph{Reproach} et publie une liste d'intentions (dans cet exemple, l'intention prioritaire est l'intention émotionnelle d'exprimer le reproche) ;
\item  Le n{\oe}ud \textit{planification} reçoit cette liste d'intentions et choisit la plus prioritaire, et publie une liste de plans (ici, un plan unique contenant l'acte ACM \emph{Reproach}).
\item Le plan est reçu par le \textit{scheduler} qui choisit la première action (ici, l'acte ACM \emph{Reproach}) ;
\item Finalement le \textit{moteur de rendu} "joue" l'acte de conversation \emph{Reproach} de façon multimodale (verbale et non-verbale) sur le robot Nao.
\end{enumerate}


\subsection{Évaluation conceptuelle}

Nous proposons ici une courte évaluation conceptuelle de l'architecture CAIO en suivant les critères d'évaluation des architectures cognitives proposés par Langley \etal \cite{Langley2009a}:
\begin{itemize}
\item \textbf{Généralisable \& Polyvalent} (facilement adaptable à différents domaines et indépendant de la tâche): la mémoire à long-terme de CAIO contient une base de connaissances génériques et statiques, et un nouveau domaine de connaissance peut être chargé dynamiquement afin de s'adapter au contexte.
\item \textbf{Rationnel \& Optimal} (prise de décision logique, sélection de comportement optimal pour un contexte donné): le raisonnement produit par l'architecture CAIO s'appuie sur 5 états mentaux et est dans la lignée des raisonnements des agents rationnels BDI, assurant que le plan choisi par le robot déduit de son intention est optimal dans le contexte actuel de sa base de connaissances.
\item \textbf{Efficace \& Extensible} (réponse et action en temps réel): l'implémentation ROS de CAIO permet une interaction humain-robot en temps réel.
\item \textbf{Réactive \& Pérenne} (réponse à des stimuli tout en gardant des buts à long terme): CAIO gère à la fois des intentions réactives à court-terme (intentions émotionnelles et de discours) et des intentions pérennes à long-terme (intentions globales).
\item \textbf{Améliorable} (comportements améliorables dans le temps, soit par programmation soit par apprentissage automatique): pour l'instant, CAIO  est améliorable seulement par programmation (chaque module peut être remplacé ou amélioré séparément), mais l'apprentissage automatique de nouvelles règles sera possible dans des extensions futures.

\item \textbf{Autonome \& Capable d'opérations prolongées} (choix de ses propres buts): CAIO permet à un robot de choisir ses propres intentions sur la base de ses perceptions, de ses états mentaux et de sa base de connaissances.
\end{itemize}

Cette rapide évaluation conceptuelle montre que l'architecture cognitive et affective CAIO possède la plupart des propriétés identifiées comme indispensables à toutes les architectures cognitives. L'architecture CAIO nécessite toutefois d'être expérimentée dans différents scénarii d'interaction Humain-Robot. Elle nécessite d'être améliorée au niveau de la représentation des connaissances (utilisation d'un langage de représentation des connaissances comme RDF ou OWL) et en se dotant de capacité d'apprentissage.




\section{Conclusion}        
\label{conclu}              

Dans cet article, nous avons décrit l'architecture Cognitive et Affective Orientée Interaction CAIO pour robot social. Cette architecture est en cours de validation plus poussée par des interactions en temps réel avec des enfants afin de vérifier que le robot exprime clairement ses intentions, et est perçu comme sincère. Nous étudierons les limites du système sur différents types de tâches.
Des travaux sur la représentation des connaissances sont nécessaires afin d'être compatible avec les ontolgies web. Le module de perception multimodale doit également être complété pour que l'extraction de l'acte de conversation ACM de l'énoncé de l'utilisateur soit améliorée, mais aussi pour y intégrer d'autres modalités (reconnaissance faciale des émotions, reconnaissance de gestes par exemple).
Un module d'apprentissage serait aussi une extension intéressante pour s'assurer que le robot s'améliore durant l'interaction, et apprenne progressivement à connaître son utilisateur, à s'y adapter et à l'engager dans l'interaction.

\enlargethispage{30pt}

\end{document}